\documentclass[11pt,a4paper]{article}
\usepackage[hyperref]{emnlp-ijcnlp-2019}
\usepackage{times}
\usepackage{latexsym,tabularx,makecell}
\usepackage{algorithm,algcompatible,amsmath}
\usepackage{url}
\usepackage{dirtytalk}
\usepackage{todonotes}

\usepackage{mathtools} 
\usepackage{esvect} 
\usepackage{amssymb}
\usepackage{bm}
\usepackage[inline]{enumitem}
\usepackage{array,colortbl,multirow,multicol,booktabs,ctable}
\aclfinalcopy 
\def\aclpaperid{3334} 
 
\title{Neural Extractive Text Summarization with Syntactic Compression}

\author{Jiacheng Xu ~{\normalfont and}~ Greg Durrett \\
 Department of Computer Science \\
 The University of Texas at Austin\\
 {\tt \{jcxu,gdurrett\}@cs.utexas.edu}}
\date{}

\begin{document}
\maketitle
\begin{abstract}
Recent neural network approaches to summarization are largely either selection-based extraction or generation-based abstraction. In this work, we present a neural model for single-document summarization based on joint extraction and syntactic compression. Our model chooses sentences from the document, identifies possible compressions based on constituency parses, and scores those compressions with a neural model to produce the final summary. For learning, we construct oracle extractive-compressive summaries, then learn both of our components jointly with this supervision. Experimental results on the CNN/Daily Mail and New York Times datasets show that our model achieves strong performance (comparable to state-of-the-art systems) as evaluated by ROUGE. Moreover, our approach outperforms an off-the-shelf compression module, and human and manual evaluation shows that our model's output generally remains grammatical.


\end{abstract}

\section{Introduction}


Neural network approaches to document summarization have ranged from purely extractive \cite{Cheng_Neural_2016,Nallapati_SummaRuNNer_2017,Narayan_Ranking_2018} to abstractive \citep{Rush_A_2015,Nallapati_Abstractive_2016,Chopra_Abstractive_2016,Tan_Abstractive_2017,Gehrmann_Bottom_2018}. Extractive systems are robust and straightforward to use. Abstractive systems are more flexible for varied summarization situations \cite{GruskyEtAl2018}, but can make factual errors \cite{CaoEtAl2018fact,LiEtAl2018} or fall back on extraction in practice \citep{See_Get_2017}. Extractive and compressive systems \cite{Berg-Kirkpatrick_Jointly_2011,Qian_Fast_2013,Durrett_Learning_2016} combine the strengths of both approaches; however, there has been little work studying neural network models in this vein, and the approaches that have been employed typically use seq2seq-based sentence compression \cite{Chen_Fast_2018}.


In this work, we propose a model that can combine the high performance of neural extractive systems, additional flexibility from compression, and interpretability given by having discrete compression options. Our model first encodes the source document and its sentences and then sequentially selects a set of sentences to further compress. Each sentence has a set of compression options available that are selected to preserve meaning and grammaticality; these are derived from syntactic constituency parses and represent an expanded set of discrete options from prior work \cite{Berg-Kirkpatrick_Jointly_2011,Wang_A_2013}. The neural model additionally scores and chooses which compressions to apply given the context of the document, the sentence, and the decoder model's recurrent state.





A principal challenge of training an extractive and compressive model is constructing the oracle summary for supervision. We identify a set of high-quality sentences from the document with beam search and derive oracle compression labels in each sentence through an additional refinement process. Our model's training objective combines these extractive and compressive components and learns them jointly.

\begin{figure}[t!]
\centering
\includegraphics[width=0.48125\textwidth]{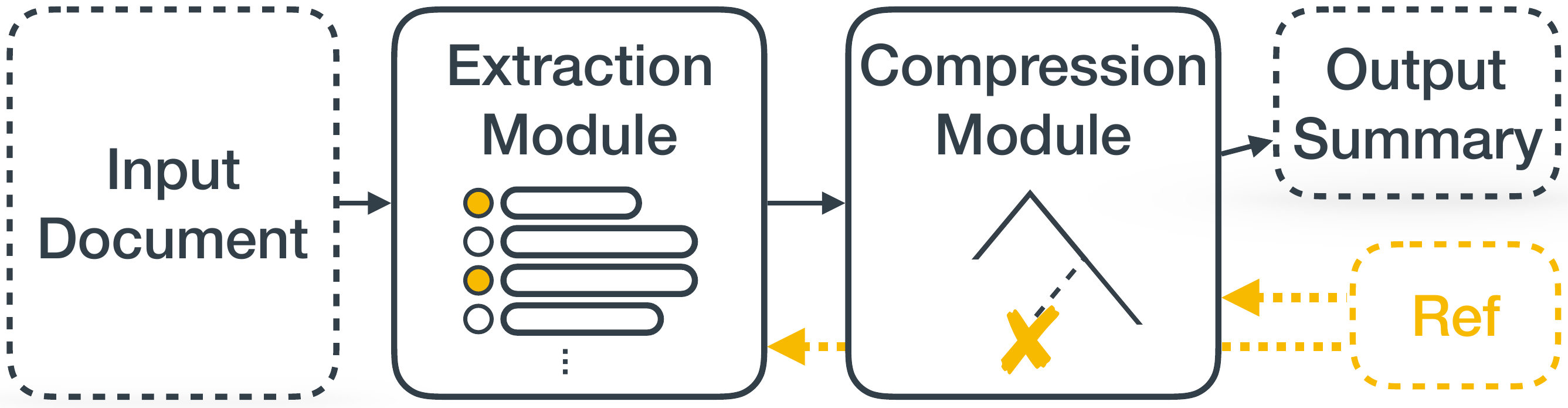}
\caption{ Diagram of the proposed model. Extraction and compression are modularized but jointly trained with supervision derived from the reference summary.
} 
\label{fig:diagram}
\end{figure}

We conduct experiments on standard single document news summarization datasets: CNN, Daily Mail \cite{Hermann_Teaching_2015}, and the New York Times Annotated Corpus \cite{Sandhaus_The_2008}. 
Our model matches or exceeds the state-of-the-art on all of these datasets and achieves the largest improvement on CNN (+2.4 ROUGE-F$_1$ over our extractive baseline) due to the more compressed nature of CNN summaries. We show that our model's compression threshold is robust across a range of settings yet tunable to give different-length summaries. Finally, we investigate the fluency and grammaticality of our compressed sentences. The human evaluation shows that our system yields generally grammatical output, with many remaining errors being attributed to the parser.\footnote{The code, full model output, and the pre-trained model are available at \url{https://github.com/jiacheng-xu/neu-compression-sum}}


\section{ Compression in Summarization}
\label{sec:bg}



Sentence compression is a long-studied problem dealing with how to delete the least critical information in a sentence to make it shorter \citep{Knight_Statistics_2000,Knight_Summarization_2002,Martins_Summarization_2009,Cohn_Sentence_2009,Wang_A_2013,Li_Improving_2014}. Many of these approaches are syntax-driven, though end-to-end neural models have been proposed as well \cite{Filippova_Sentence_2015,wang-etal-2017-syntax}. Past non-neural work on summarization has used both syntax-based \cite{Berg-Kirkpatrick_Jointly_2011,Woodsend_Learning_2011} and discourse-based \cite{Carlson_Building_2001,Hirao_Single_2013,Li_The_2016} compressions. Our approach follows in the syntax-driven vein.

\begin{figure}[t]
\centering
\includegraphics[width=0.48125\textwidth]{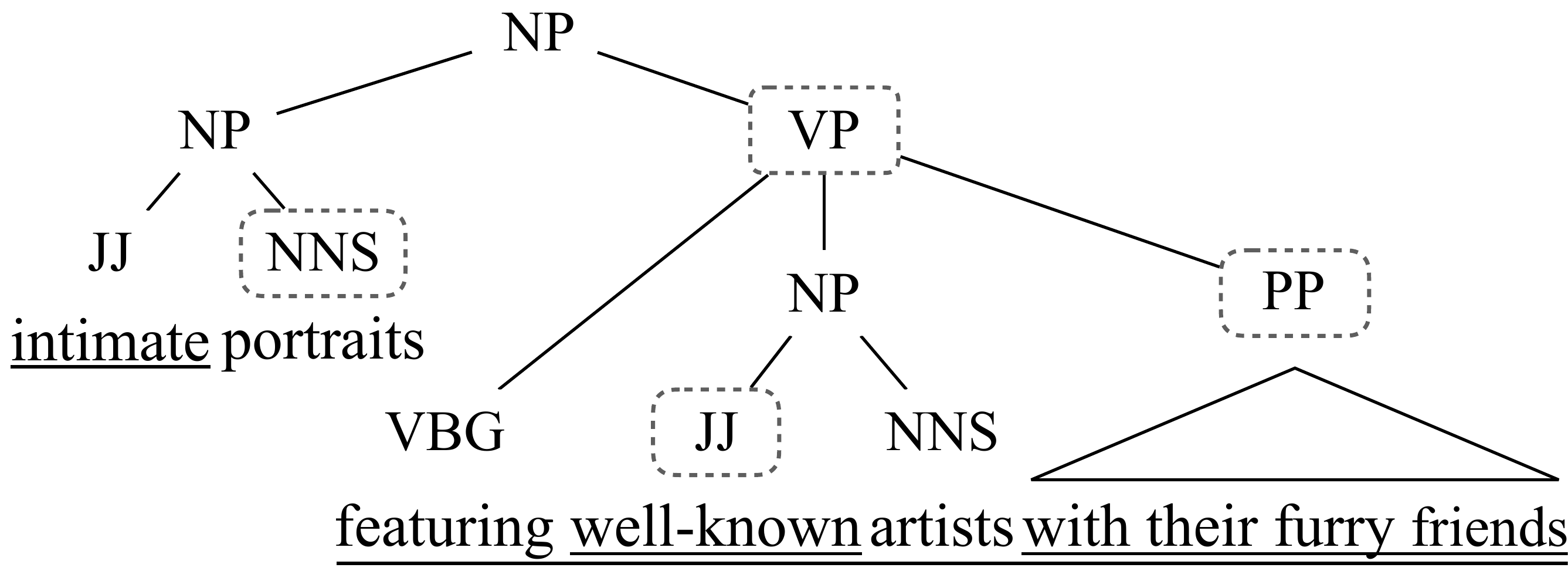}
\caption{ Text compression example. In this case, \say{intimate}, \say{well-known}, \say{with their furry friends} and \say{featuring ... friends} are deletable given compression rules.
} 
\label{fig:example}
\end{figure}




\begin{figure*}[t!]
\centering
\includegraphics[width=1\textwidth]{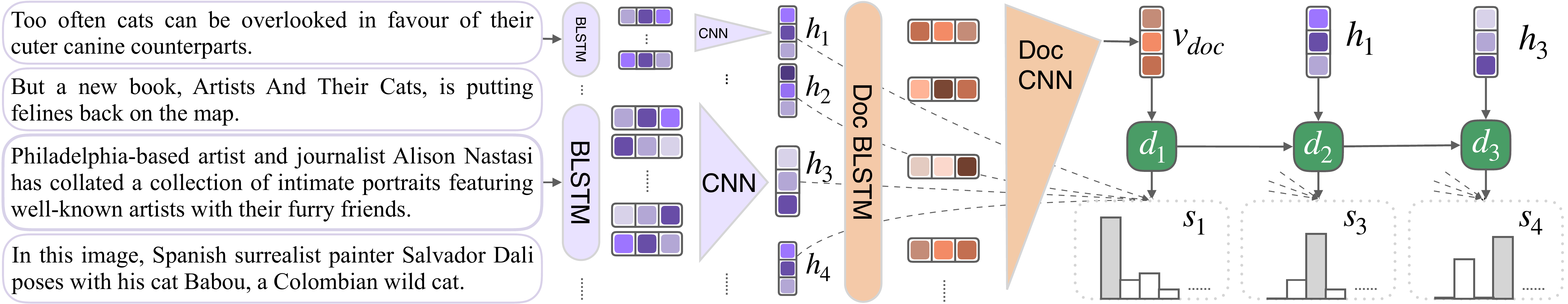}
\caption{Sentence extraction module of \textsc{JECS}. Words in input document sentences are encoded with BiLSTMs. Two layers of CNNs aggregate these into sentence representations $h_i$ and then the document representation $v_{doc}$. This is fed into an attentive LSTM decoder which selects sentences based on the decoder state $d$ and the representations $h_i$, similar to a pointer network.}
\label{fg:model}
\end{figure*}


Our high-level approach to summarization is shown in Figure~\ref{fig:diagram}. In Section~\ref{sec:model}, we describe the models for extraction and compression. Our compression depends on having a discrete set of valid compression options that maintain the grammaticality of the underlying sentence, which we now proceed to describe.

\paragraph{Compression Rules}
\label{para:rules}
We refer to the rules derived in \citet{Li_Improving_2014}, \citet{Wang_A_2013}, and \citet{Durrett_Learning_2016} and design a concise set of syntactic rules including the removal of: \begin{enumerate*}
    \item Appositive noun phrases;
    \item Relative clauses and adverbial clauses;
    \item Adjective phrases in noun phrases, and adverbial phrases (see Figure~\ref{fig:example});
    \item Gerundive verb phrases as part of noun phrases (see Figure~\ref{fig:example});
    \item Prepositional phrases in certain configurations like \emph{on Monday};
    \item Content within parentheses and other parentheticals.
\end{enumerate*}

Figure~\ref{fig:example} shows examples of several compression rules applied to a short snippet. All combinations of compressions maintain grammaticality, though some content is fairly important in this context (the VP and PP) and should not be deleted. Our model must learn not to delete these elements.

\paragraph{Compressability} Summaries from different sources may feature various levels of compression. At one extreme, a summary could be fully sentence-extractive; at another extreme, the editor may have compressed a lot of content in a sentence. In Section~\ref{sec:train}, we examine this question on our summarization datasets and use it to motivate our choice of evaluation datasets.

\paragraph{Universal Compression with ROUGE} While we use syntax as a source of compression options, we note that other ways of generating compression options are possible, including using labeled compression data. However, supervising compression with ROUGE is critical to learn what information is important for this particular source, and in any case, labeled compression data is unavailable in many domains. In Section~\ref{sec:experiments}, we compare our model to off-the-shelf sentence compression module and find that it substantially underperforms our approach.

\section{Model}
\label{sec:model}

\begin{figure}[t!]
\centering
\includegraphics[width=0.48125\textwidth]{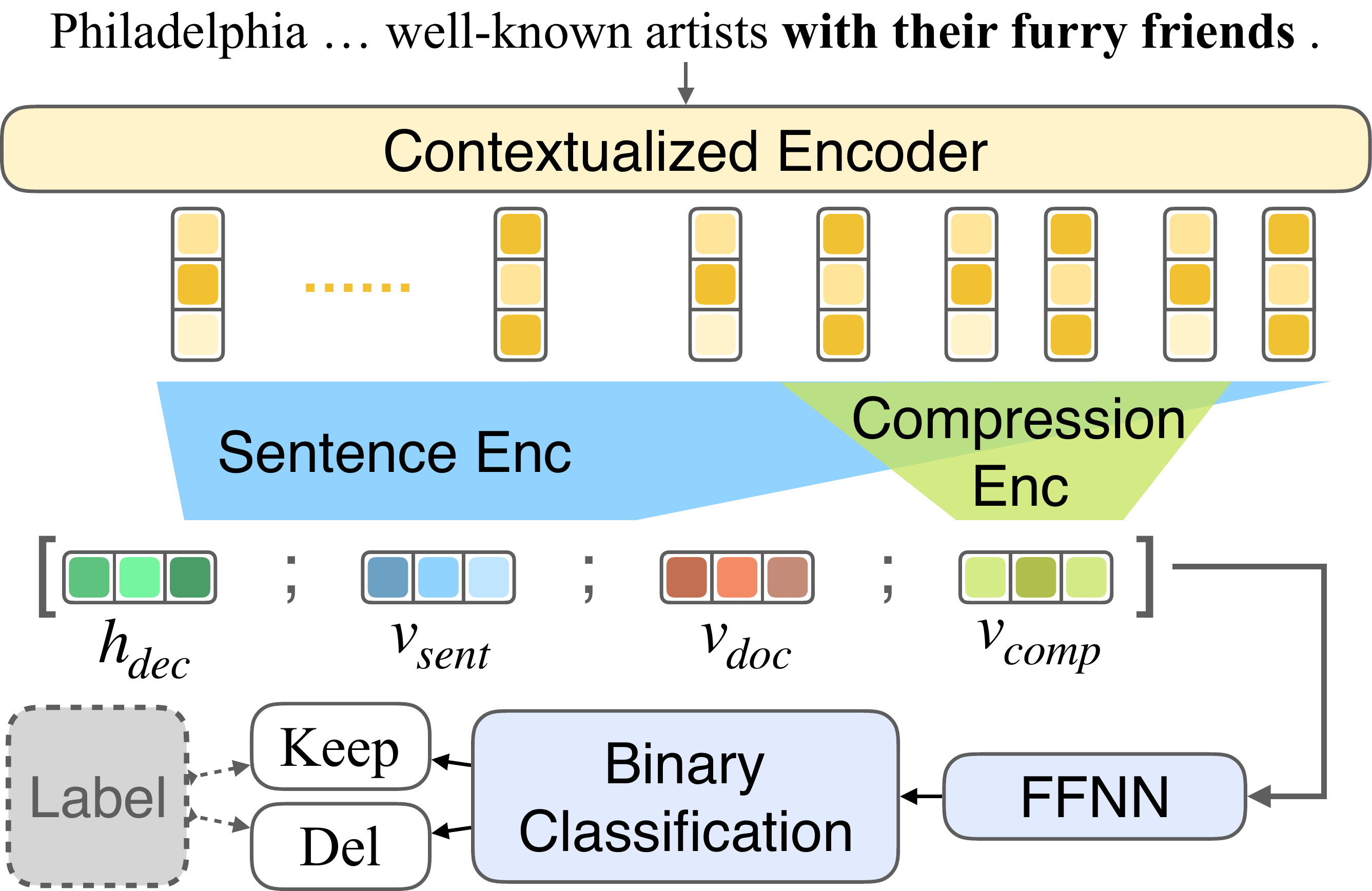}
\caption{Text compression module. A neural classifier scores the compression option (\emph{with their furry friends}) in the sentence and broader document context and decides whether or not to delete it.}
\label{fg:cp}
\end{figure}

Our model is a neural network model that encodes a source document, chooses sentences from that document, and selects discrete compression options to apply.
The model architecture of sentence extraction module and text compression module are shown in Figure~\ref{fg:model} and \ref{fg:cp}. 


\subsection{Extractive Sentence Selection}

A single document consists of $n$ sentences $D= \{s_1, s_2, \cdots , s_n\}$. The $i$-th sentence is denoted as $s_i=\{w_{i1}, w_{i2}, \cdots ,w_{im}\}$ where $w_{ij}$ is the $j$-th word in $s_i$. The content selection module learns to pick up a subset of $D$ denoted as $\hat{D} = \{ \hat{s}_1, \hat{s}_2,  \cdots , \hat{s}_k, | \hat{s}_i \in D \}$ where $k$ sentences are selected.

\paragraph{Sentence \& Document Encoder}
We first use a bidirectional LSTM to encode words in each sentence in the document separately and then we apply multiple convolution layers and max pooling layers to extract the representation of every sentence. Specifically, $[h_{i1}, \cdots, h_{im}] = \text{BiLSTM}([w_{i1},  \cdots ,w_{im}])$ and $h_{i} = \text{CNN}([h_{i1}, \cdots, h_{im}])$
where $h_i$ is a representation of the $i$-th sentence in the document. This process is shown in the left side of Figure~\ref{fg:model} illustrated in purple blocks.
We then aggregate these sentence representations into a document representation $v_{doc}$ with a similar BiLSTM and CNN combination, shown in Figure~\ref{fg:model} with orange blocks.

\paragraph{Decoding}
The decoding stage selects a number of sentences given the document representation $v_{doc}$ and sentences' representations $h_i$. This process is depicted in the right half of Figure~\ref{fg:model}. We use a sequential LSTM decoder where, at each time step, we take the representation $h$ of the last selected sentence, the overall document vector $v_{doc}$, and the recurrent state $d_{t-1}$, and produce a distribution over all of the remaining sentences excluding those already selected. This approach resembles pointer network-style approaches used in past work \cite{Zhou_Neural_2018}.
Formally, we write this as:
\begin{align*}
    d_t = \text{LSTM}( d_{t-1}, h_{k}, v_{doc})\\
   \text{score}_{t,i} = W_m \text{tanh} (W_d d_t + W_h h_i)\\
   p(\hat{s}_t=s_i|d_t,h_{k},v_{doc},h_i)= \text{softmax}(\text{score}_{t,i})
\end{align*}
where $ h_{k}$ is the representation of the sentence selected at time step $t-1$. $d_{t-1}$ is the decoding hidden state from last time step. $W_d$, $W_h$, $W_m$, and parameters in LSTM are learned. Once a sentence is selected, it cannot be selected again. At test time, we use greedy decoding to identify the most likely sequence of sentences under our model.\footnote{For our experiments, we decode for a fixed number of sentences, tuned for each dataset, as in prior extractive work \cite{Narayan_Ranking_2018}. We experimented with dynamically choosing a number of sentences and found this to make little difference.}



\subsection{Text Compression}

After selecting the sentences, the text compression module evaluates our discrete compression options and decides whether to remove certain phrases or words in the selected sentences. Figure~\ref{fg:cp} shows an example of this process for deciding whether or not to delete a PP in this sentence. This PP was marked as deletable based on rules described in Section~\ref{sec:bg}. Our network then encodes this sentence and the compression, combines this information with the document context $v_{doc}$ and decoding context $h_{dec}$, and uses a feedforward network to decide whether or not to delete the span.
Let $C_i=\{ c_{i1}, \cdots , c_{il} \}$ denote the possible compression spans derived from the rules described in Section~\ref{para:rules}. Let $y_{i,c}$ be a binary variable equal to 1 if we are deleting the $c$th option of the $i$th sentence. Our text compression module models $p(y_{i,c}|D,\hat{s}_t = s_i)$ as described in the following section.



\paragraph{Compression Encoder} We use a contextualized encoder, ELMo \citep{Peters_Deep_2018} to compute contextualized word representations. We then use CNNs with max pooling to encode the sentence (shown in blue in Figure~\ref{fg:cp}) and the candidate compression (shown in light green in Figure~\ref{fg:cp}). The sentence representation $v_{sent}$ and the compression span representation $v_{comp}$ are concatenated with the hidden state in sentence decoder $h_{dec}$ and the document representation $v_{doc}$.

\paragraph{Compression Classifier}
We feed the concatenated representation to a feedforward neural network to predict whether the compression span should be deleted or kept, which is formulated as a binary classification problem. This classifier computes the final probability $p(y_{i,c}|D,\hat{s}_t = s_i) = p(y_{i,c}|h_{dec},v_{doc},v_{comp},s_i)$.
The overall probability of a summary $(\hat{s},\hat{y})$, where $\hat{s}$ is the sentence oracle and $\hat{y}$ is the compression label, is the product of extraction and compression models: $p(\hat{s},\hat{y}|D) = \prod_{t=1}^T \left[ p(\hat{s}_t |D, \hat{s}_{<t})  \prod_{c \in C_i} p(\hat{y}_{t,c} | D,\hat{s}) \right]$.



\paragraph{Heuristic Deduplication}
Inspired by the trigram avoidance trick proposed in \citet{Paulus_A_2018} to reduce redundancy, we take full advantage of our linguistically motivated compression rules and the constituent parse tree and allow our model to compress deletable chunks with redundant information. 
We therefore take our model's output and apply a postprocessing stage where we remove any compression option whose unigrams are completely covered elsewhere in the summary. We perform this compression after the model prediction and compression.

\section{Training}
\label{sec:train}

Our model makes a series of sentence extraction decisions $\hat{s}$ and then compression decisions $\hat{y}$. To supervise it, we need to derive gold-standard labels for these decisions. Our oracle identification approach relies on first identifying an oracle set of sentences and then the oracle compression options.\footnote{Depending on which sentences are extracted, different compression decisions may be optimal; however, re-deriving these with a dynamic oracle \cite{GoldbergNivre2012} is prohibitively expensive during training.}

\subsection{Oracle Construction}
\label{sec:oracle}

\begin{table}[]
\small
\begin{tabular}{l ccc}
\multicolumn{4}{p{0.46\textwidth}}{ \textbf{Reference}: Artist and journalist Alison Nastasi put together the portrait collection. Also features images of Picasso, Frida Kahlo, and John Lennon. Reveals quaint personality traits shared between artists and their felines.} \\
\multicolumn{4}{p{0.46\textwidth}}{\textbf{Document}: ... \textbf{Philadelphia-based} artist and journalist Alison Nastasi has collated a collection of \textbf{intimate} portraits \textbf{featuring well-known artists with their furry friends}. ...}       \\
\toprule
\multirow{1}{*}{Compression  $R_{bf}=19.4$}  &  $R_{af}$   & Ratio    & Label\\\midrule
Philadelphia-based & 19.8  & 1.02 & DEL \\ 
intimate  & 19.8 & 1.02 & DEL \\  
well known  & 20.4 & 1.05 & DEL\\  
featuring ... their furry friends   & 18.1                   & 0.93                  & KEEP                 \\ \bottomrule
\end{tabular}
\caption{Oracle label computation for the text compression module. $R_{bf}$ and  $R_{af}$  are the ROUGE scores before and after compression. The ratio is defined as $\frac{R_{af}}{R_{bf}}$. ROUGE increases when words not appearing in the reference are deleted. ROUGE can decrease when terms appearing in the reference summary, like \emph{featuring}, are deleted.
}
\end{table}

\paragraph{Sentence Extractive Oracle} We first identify an oracle set of sentences to extract using a beam search procedure similar to Maximal Marginal Relevance (MMR) \cite{Carbonell_The_1998}. For each additional sentence we propose to add, we compute a heuristic cost equal to the ROUGE score of a given sentence with respect to the reference summary. When pruning states, we calculate the ROUGE score of the combination of sentences currently selected and sort in descending order. 
Let the beam width be $\beta$. The time complexity of the approximate approach is $O(nk\beta )$ where in practice $k \ll n$ and $\beta \ll n$. We set $\beta=8$ and $n=30$ which means we only consider the first 30 sentences in the document.

The beam search procedure returns a beam of $\beta$ different sentence combinations in the final beam. 
We use the sentence extractive oracle for both the extraction-only model and the joint extraction-compression model.

\begin{table}[t]
  \centering
  \small
  \begin{tabular}{c|ccc}
  \toprule
  \multirow{1}{*}{Category} 
     & CNN & DM & NYT   \\
\midrule
Bad	&	27\%	&	48\%	&	49\% \\
Weak Positive	&	58\%	&	43\%	&	47\% \\
Strong Positive	&	15\%	&	10\%	&	4\% \\
\bottomrule
  \end{tabular}
  \caption{Compressibility: The oracle label distribution over three datasets. Compressions in the \say{Bad} category decrease ROUGE and are labeled as negative (do not delete), while weak positive (less than 5\% ROUGE improvement) and strong positive (greater than 5\%) both represent ROUGE improvements. CNN features much more compression than the other datasets.
  }
  \label{tb:dataset_compare}
\end{table}


\paragraph{Oracle Compression Labels} To form our joint extractive and compressive oracle, we need to give the compression decisions binary labels $y_{i,c}$ in each set of extracted sentences. 
For simplicity and computational efficiency, we assign each sentence a single $y_{i,c}$ independent of the context it occurs in.
For each compression option, we assess the value of it by comparing the ROUGE score of the sentence with and without this phrase. Any option that increases ROUGE is treated as a compression that should be applied. When calculating this ROUGE value, we remove stop words include stemming.

We run this procedure on each of our oracle extractive sentences. The fraction of positive and negative labels assigned to compression options is shown for each of the three datasets in Table \ref{tb:dataset_compare}. CNN is the most compressable dataset among CNN, DM and NYT.


\paragraph{ILP-based oracle construction} Past work has derived oracles for extractive and compressive systems using integer linear programming (ILP) \cite{Gillick_A_2009,Berg-Kirkpatrick_Jointly_2011}. Following their approach, we can directly optimize for ROUGE recall of an extractive or compressive summary in our framework if we specify a length limit. However, we evaluate on ROUGE F$_1$ as is standard when comparing to neural models that don't produce fixed-length summaries. Optimizing for ROUGE F$_1$ cannot be formulated as an ILP, since computing precision requires dividing by the number of selected words, making the objective no longer linear. 
We experimented with optimizing for ROUGE F$_1$ indirectly by finding optimal ROUGE recall summaries at various settings of maximum summary length. However, these summaries frequently contained short sentences to fill up the budget, and the collection of summaries returned tended to be less diverse than those found by beam search.

\subsection{Learning Objective}
\label{sec:learning}

Often, many oracle summaries achieve very similar ROUGE values. We therefore want to avoid committing to a single oracle summary for the learning process. Our procedure from Section~\ref{sec:oracle} can generate $m$ extractive oracles $s^{*}_{i}$; let $s^{*}_{i,t}$ denote the gold sentence for the $i$-th oracle at timestep $t$. Past work \cite{Narayan_Ranking_2018,Chen_Fast_2018} has employed policy gradient in this setting to optimize directly for ROUGE. However, because oracle summaries usually have very similar ROUGE scores, we choose to simplify this objective as $\mathcal{L}_{\text{sent}}= - \frac{1}{m} \sum_{i=1}^{m} \sum_{t=1}^{T} \log{p(s^{*}_{i,t}|D,s^{*}_{i,<t})  } $.
Put another way, we optimize the log likelihood averaged across $m$ different oracles to ensure that each has high likelihood.
We use $m=5$ oracles during training. The oracle sentence indices are sorted according to the individual salience (ROUGE score) rather than document order.

The objective of the compression module is defined as $ \mathcal{L}_{\text{comp}}= - \sum_{i=1}^{m} \sum_{c=1}^{C} \log{p(y^{*}_{i,c}|D,\hat{s})  }  $ where $p(y^{*}_{i,c})$ is the probability of the target decision for the $c$-th compression options of the $i$-th sentence. 
The joint loss function is $\mathcal{L}=\mathcal{L}_{\text{sent}}+\alpha \mathcal{L}_{\text{comp}}$. We set $\alpha=1$ in practice.








\section{Experiments}
\label{sec:experiments}

We evaluate our model on two axes. First, for content selection, we use ROUGE as is standard.
Second, we evaluate the grammaticality of our model to ensure that it is not substantially damaged by compression. 

\subsection{Experimental Setup}
\paragraph{Datasets} We evaluate the proposed method on three popular news summarization datasets: the New York Times corpus \cite{Sandhaus_The_2008}, CNN and Dailymail (DM) \cite{Hermann_Teaching_2015}.\footnote{More details about the experimental setup, implementation details, and human evaluation are provided in the Appendix. \nocite{Manning_The_2014,Kingma_Adam_2014}}

As discussed in Section~\ref{sec:bg}, compression will give different results on different datasets depending on how much compression is optimal from the standpoint of reproducing the reference summaries, which changes how measurable the impact of compression is. In Table~\ref{tb:dataset_compare}, we show the ``compressability'' of these three datasets: how valuable various compression options seem to be from the standpoint of improving ROUGE. We found that CNN has significantly more positive compression options than the other two. Critically, CNN also has the shortest references (37 words on average, compared to 61 for Daily Mail; see Appendix).
In our experiments, we first focus on CNN and then evaluate on the other datasets.

\paragraph{Models}
We present several variants of our model to show how extraction and compression work jointly. In extractive summarization, the \textsc{Lead} baseline (first $k$ sentences) is a strong baseline due to how newswire articles are written.
\textsc{LeadDedup} is a non-learned baseline that uses our heuristic deduplication technique on the lead sentences.
\textsc{LeadComp} is a compression-only model where compression is performed on the lead sentences. This shows the effectiveness of the compression module in isolation rather than in the context of abstraction.
\textsc{Extraction} is the extraction only model. \textsc{JECS} is the full Joint Extractive and Compressive Summarizer.

We compare our model with various abstractive and extractive summarization models. NeuSum \citep{Zhou_Neural_2018} uses a seq2seq model to predict a sequence of sentences indices to be picked up from the document. Our extractive approach is most similar to this model. 
Refresh \citep{Narayan_Ranking_2018}, BanditSum \citep{Dong_BanditSum_2018} and LatSum \citep{Zhang_Neural_2018} are extractive summarization models for comparison.
We also compare with some abstractive models including PointGenCov \citep{See_Get_2017}, FARS \citep{Chen_Fast_2018} and CBDec \citep{Jiang_Closed_2018}.

\begin{table}[t!]
\centering
\small


\begin{tabular}{l|ccc}
\toprule
\multicolumn{1}{c|}{\multirow{2}{*}{Model}} & \multicolumn{3}{c}{CNN}                                                     \\
\multicolumn{1}{c|}{}                       & \multicolumn{1}{c}{R-1} & \multicolumn{1}{c}{R-2} & \multicolumn{1}{c}{R-L} \\
\midrule
Lead (Ours)                                & 29.1                    & 11.1                    & 25.8                    \\
Refresh* \citep{Narayan_Ranking_2018}           &     30.3                    &            \textbf{ 11.6}           &        26.9                 \\
LatSum*  \citep{Zhang_Neural_2018}               &    28.8                     &             11.5            & 25.4                        \\
BanditSum      \citep{Dong_BanditSum_2018}                               &    \textbf{30.7}                     &    \textbf{11.6}                     &     \textbf{27.4}                    \\ \midrule
   \textsc{LeadDedup}    &  29.7                       &    10.9                     &        26.2                 \\
\textsc{LeadComp}          & 30.6&10.8                         &    27.2                     \\
\textsc{Extraction}                                 &      30.3                   & 11.0         &       26.5                  \\
\textsc{ExtLSTMDel}                                 &      30.6                   &11.9         &       27.1                  \\
\textsc{JECS}                                       & \textbf{32.7}                    & \textbf{12.2}                    & \textbf{29.0}           \\ \bottomrule
\end{tabular}
\caption{Experimental results on the test sets of CNN. * indicates models evaluates with our own ROUGE metrics. Our model outperforms our extractive model and lead-based baselines, as well as prior work. 
}
\label{tb:cnn}
\end{table}

We also compare our joint model with a pipeline model with an off-the-shelf compression module. We implement a deletion-based BiLSTM model for sentence compression \citep{wang-etal-2017-syntax} and run the model on top of our extraction output.\footnote{We reimplemented the authors' model following their specification and matched their accuracy. For fair comparison, we tuned the deletion threshold to match the compression rate of our model; other choices did not lead to better ROUGE scores.} The pipeline model is denoted as \textsc{ExtLstmDel}.

\begin{table}[t!]
\centering
\small


\begin{tabular}{l|ccc}
\toprule
\multicolumn{1}{c|}{\multirow{2}{*}{Model}} & \multicolumn{3}{c}{CNNDM}                                                     \\
\multicolumn{1}{c|}{}                       & \multicolumn{1}{c}{R-1} & \multicolumn{1}{c}{R-2} & \multicolumn{1}{c}{R-L} \\
\midrule
Lead  (Ours)         & 40.3 & 17.6 & 36.4       \\
Refresh*     \citep{Narayan_Ranking_2018}              & 40.0 & 18.1 & 36.6     \\
NeuSum  & \textbf{41.6} & \textbf{19.0} & 38.0          \\
LatSum*  \citep{Zhang_Neural_2018}      & 41.0 & 18.8 & 37.4           \\
LatSum w/ Compression                                                  & 36.7 & 15.4 & 34.3        \\
BanditSum                                                  & 41.5 & 18.7 & 37.6      \\
CBDec \citep{Jiang_Closed_2018}                                           & 40.7 & 17.9 & 37.1               \\
 FARS \citep{Chen_Fast_2018}      & 40.9 & 17.8 & \textbf{38.5}            \\
\midrule
\textsc{LeadDedup}                                      & 40.5 & 17.4 & 36.5         \\
\textsc{LeadComp}                                                & 40.8 & 17.4 & 36.8         \\
\textsc{Extraction}                                               & 40.7 & 18.0 & 36.8      \\
\textsc{JECS}           & \textbf{41.7} & \textbf{18.5} &\textbf{ 37.9 }      \\
\bottomrule
\end{tabular}
\caption{Experimental results on the test sets of CNNDM. 
The portion of CNN is roughly one of tenth of DM.
Gains are more pronounced on CNN because this dataset features shorter, more compressed reference summaries.
}
\label{tb:cnndm}
\end{table}

\subsection{Results on CNN}

Table~\ref{tb:cnn} shows experiments results on CNN. We list performance of the \textsc{Lead} baseline and the performance of competitor models on these datasets. Starred models are evaluated according to our ROUGE metrics; numbers very closely match the originally reported results.

Our model achieves substantially higher performance than all baselines and past systems (+2 ROUGE F1 compared to any of these). 
On this dataset, compression is substantially useful. Compression is somewhat effective in isolation, as shown by the performance of \textsc{LeadDedup} and \textsc{LeadComp}. But compression in isolation still gives less benefit (on top of \textsc{Lead}) than when combined with the extractive model (\textsc{JECS}) in the joint framework. Furthermore, our model beats the pipeline model \textsc{ExtLSTMDel} which shows the necessity of training a joint model with ROUGE supervision.

\subsection{Results on Combined CNNDM and NYT}
We also report the results on the full CNNDM and NYT although they are less compressable. Table~\ref{tb:cnndm} and Table~\ref{tb:nyt} shows the experimental results on these datasets.


Our models still yield strong performance compared to baselines and past work on the CNNDM dataset. The \textsc{Extraction} model achieves comparable results to past successful extractive approaches on CNNDM and \textsc{JECS} improves on this across the datasets. 
In some cases, our model slightly underperforms on ROUGE-2. One possible reason is that we remove stop words when constructing our oracles, which could underestimate the importance of bigrams containing stopwords for evaluation.
Finally, we note that our compressive approach substantially outperforms the compression-augmented LatSum model. That model used a separate seq2seq model for rewriting, which is potentially harder to learn than our compression model.

\begin{table}[t]
  \centering
  \small
  \begin{tabular}{l|ccc}
  \toprule
  \multirow{1}{*}{Model} 
     & R-1 & R-2 & R-L   \\
\midrule
Lead & 41.8	&	22.6	&	35.0   \\
\textsc{LeadDedup}  &42.0	&	22.8	&	35.0   \\
\textsc{LeadComp}  & 42.4	&	22.7	&	35.4   \\
\textsc{Extraction}  & 44.3	&	25.5	&	37.1 \\
\textsc{JECS}  & 45.5	&	25.3	&	38.2  \\
\bottomrule
  \end{tabular}
  \caption{Experimental results on the NYT50 dataset. ROUGE-1, -2 and -L F$_1$ is reported. \textsc{JECS} substantially outperforms our Lead-based systems and our extractive model.}
  \label{tb:nyt}
\end{table}

On NYT, we see again that the inclusion of compression leads to improvements in both the \textsc{Lead} setting as well as for our full \textsc{JECS} model.\footnote{\citet{Paulus_A_2018} do not use the NYT50 dataset, so our results are not directly comparable to theirs. \citet{Durrett_Learning_2016} use a different evaluation setup with a hard limit on the summary length and evaluation on recall only.}

\subsection{Grammaticality}

We evaluate grammaticality of our compressed summaries in three ways. First, we use Amazon Mechanical Turk to compare different compression techniques. Second, to measure absolute grammaticality, we use an automated out-of-the-box tool Grammarly. Finally, we conduct manual analysis.

\begin{table}[]
  \centering
  \small
  \begin{tabular}{l|ccc}
  \toprule
  \multirow{1}{*}{Model} 
     & Preference (\%) $\uparrow$ & Error $\downarrow$ & R-1 $\uparrow$  \\
\midrule
\textsc{Ext Lead3}   &	-- & 22  & 29.1 \\
\textsc{ExtDrop}   &	12\% & 161  & 30.2 \\
\textsc{ExtLstmDel} &	43\% & \textbf{24} &   30.6\\
\textsc{JECS}	 & \textbf{45}\% & 31 &  \textbf{32.7}\\
\bottomrule
  \end{tabular}
  \caption{Human preference, ROUGE and Grammarly grammar checking results. We asked Turkers to rank the models' output based on grammaticality. 
  Error shows the number of grammar errors in 500 sentences reported by Grammarly. Our \textsc{JECS} model achieves the highest ROUGE and is preferred by humans while still making relatively few errors.
  }
  \label{tb:human}
\end{table}

\paragraph{Human Evaluation} We first conduct a human evaluation on the Amazon Mechanical Turk platform.
We ask Turkers to rank different compression versions of a sentence in terms of grammaticality. We compare our full \textsc{JECS} model and the off-the-shelf pipeline model \textsc{ExtLstmDel}, which have matched compression ratios.
We also propose another baseline, \textsc{ExtractDropout}, which randomly drops words in a sentence to match the compression ratio of the other two models. 
The results are shown in Table~\ref{tb:human}. Turkers give roughly equal preference to our model and the \textsc{ExtLstmDel} model, which was learned from supervised compression data. However, our \textsc{JECS} model achieves substantially higher ROUGE score, indicating that it represents a more effective compression approach.

We found that absolute grammaticality judgments were hard to achieve on Mechanical Turk; Turkers' ratings of grammaticality were very noisy and they did not consistently rate true article sentences above obviously noised variants. Therefore, we turn to other methods as described in the next two paragraphs.

\paragraph{Automatic Grammar Checking} We use Grammarly to check 500 sentences sampled from the outputs of the three models mentioned above from CNN. Both \textsc{ExtLstmDel} and \textsc{JECS} make a small number of grammar errors, not much higher than the purely extractive \textsc{Lead3} baseline. One major source of errors for \textsc{JECS} is having the wrong article after the deletion of an adjective like \emph{an [awesome] style}. 

\begin{table*}[]
\centering
\small
\begin{tabular}{p{0.28\textwidth}|p{0.61\textwidth}}

\toprule
\textbf{Reference Summary} & \textbf{Prediction with \textcolor{gray}{Compressions}}              \\
\midrule
  Mullah Omar, the reclusive founder of the Afghan Taliban, is still in charge, a new biography claims. An ex-Taliban insider says there have been rumors that the one-eyed militant is dead.
  & \textcolor{gray}{(CNN)} Mullah Mohammed Omar is ``still the leader'' of the Taliban's self-declared Islamic Emirate of Afghanistan. The Taliban's ``Cultural Commission'' released the 11-page document in \textcolor{gray}{several different} translations on the movement's website, \textcolor{gray}{ostensibly} to commemorate the \textcolor{gray}{19th} anniversary \textcolor{gray}{of an April 4, 1996, meeting in Afghanistan's Kandahar province} when an assembly of Afghans swore allegiance to Omar. \\\midrule
  Rebecca Francis' photo with a giraffe was shared by Ricky Gervais. Francis was threatened on Twitter for the picture. Francis, a hunter, said the giraffe was "close to death" and became food for locals.
  &  \textcolor{gray}{(CNN)} Five years ago, Rebecca Francis posed for a photo while lying next to a dead giraffe. 
The trouble started Monday, when comedian Ricky Gervais tweeted the photo \textcolor{gray}{with a question}. 
Francis, who has appeared on the NBC Sports Network outdoor lifestyle show ``Eye of the Hunter'' and was the subject of an interview with Hunting Life \textcolor{gray}{in late March}, responded \textcolor{gray}{in a statement} to HuntingLife.com \textcolor{gray}{ on Tuesday, which was posted on its Facebook page}.   \\\midrule
Frida Ghitis: President Barack Obama is right to want a deal, but this one gives Iran too much. She says the framework agreement starts lifting Iran sanctions much too soon. & \textcolor{gray}{(CNN)} President Barack Obama tied himself to the mast of a nuclear deal with Iran even before he became the Democratic candidate for president. Reaching a \textcolor{gray}{good}, \textcolor{gray}{solid} agreement with Iran is a worthy, desirable goal. But the process has unfolded under the \textcolor{gray}{destructive} influence of political considerations, weakening America's hand and strengthening Iran. 
\\\bottomrule   \end{tabular}
\caption{Examples of applied compressions. The top two are sampled from among the most compressed examples in the dataset. Our \textsc{JECS} model is able to delete both large chunks (especially temporal PPs giving dates of events) as well as individual modifiers that aren't determined to be relevant to the summary (e.g., the specification of the \emph{19th} anniversary). The last example features more modest compression.}
\label{tb:examples}
\end{table*}

\paragraph{Manual Error Analysis} To get a better sense of our model's output, we conduct a manual analysis of our applied compressions to get a sense of how many are valid. 
We manually examined 40 model summaries, comparing the output with the raw sentences before compression, and identified the following errors: \begin{enumerate*}
  \item Eight bad deletions due to parsing errors like \emph{a UK [JJ national] from London}.
  \item Eight inappropriate adjective deletions causing correctness issues with respect to the reference document like \emph{[former] president} and \emph{[nuclear] weapon}.
  \item Three other errors: partial deletion of slang, inappropriate PP attachment deletion, and an unhandled grammatical construction: \emph{students [first], athletes [second]}.
\end{enumerate*}

Examples of output are shown in Table~\ref{tb:examples}. The first two examples are sampled from the top 25\% of the most compressed examples in the corpus. We see a variety of compression options that are used in the first two examples, including removal of temporal PPs, large subordinate clauses, adjectives, and parentheticals. The last example features less compression, only removing a handful of adjectives in a manner which slightly changes the meaning of the summary.

Improving the parser and deriving a more semantically-aware set of compression rules can help achieving better grammaticality and readability. However, we note that such errors are largely orthogonal to the core of our approach; a more refined set of compression options could be dropped into our system and used without changing our fundamental model.

\begin{figure}[t!]
\centering
\includegraphics[width=0.48\textwidth]{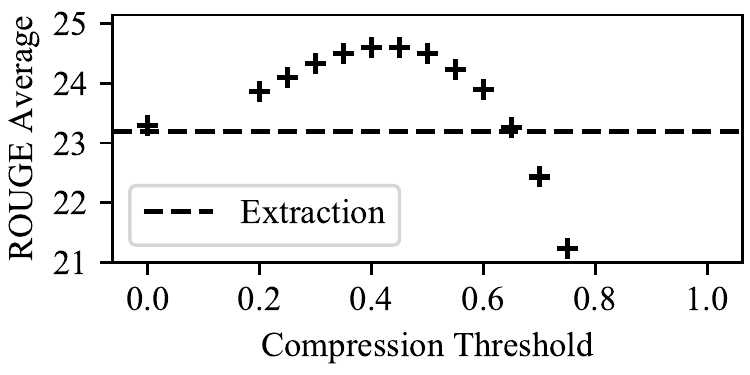}
\caption{ Effect of changing the compression threshold on CNN. The y-axis shows the average of the F1 of ROUGE-1,-2 and -L. The dotted line is the extractive baseline. The model outperforms the extractive model and achieves nearly optimal performance across a range of threshold values.} 
\label{fig:thres}
\end{figure}

\section{Compression Analysis}
\label{sec:ana}

\paragraph{Compression Threshold}
Compression in our model is an imbalanced binary classification problem. 
The trained model's natural classification threshold (probability of DEL $>$ 0.5) may not be optimal for downstream ROUGE. We experiment with varying the classification threshold from 0 (no deletion, only heuristic deduplication) to 1 (all compressible pieces removed). The results on CNN are shown in Figure~\ref{fig:thres}, where we show the average ROUGE value at different compression thresholds. The model achieves the best performance at 0.45 but performs well in a wide range from 0.3 to 0.55. Our compression is therefore robust yet also provides a controllable parameter to change the amount of compression in produced summaries.


\begin{table}[]
\centering
\small
\renewcommand{\tabcolsep}{1.8mm}
\begin{tabular}{c|cccc}\toprule
Node Type & \multicolumn{1}{l}{Len} & \multicolumn{1}{l}{\% of comps} & \multicolumn{1}{l}{Comp Acc} & \multicolumn{1}{l}{Dedup}  \\\midrule
JJ        & 1.0                           & 34\%                           & 72\%                         & 30\%                        \\
PP        & 3.4                           & 26\%                           & 47\%                         & 72\%                        \\
ADVP      & 1.4                           & 17\%                           & 79\%                         & 17\%                        \\
PRN       & 2.2                           & 6\%                            & 80\%                         & 5\%                         \\\bottomrule 
\end{tabular}
\caption{The compressions used by our model on CNN; average lengths and the fraction of that constituency type among compressions taken by our model. Comp Acc indicates how frequently that compression was taken by the oracle; note that error, especially keeping constituents that we shouldn't, may have minimal impact on summary quality.  Dedup indicates the percentage of chosen compressions which arise from deduplication as opposed to model prediction. 
}
  \label{tb:surv-af}
\end{table}

\paragraph{Compression Type Analysis} We further break down the types of compressions used in the model. Table~\ref{tb:surv-af} shows the compressions that our model ends up choosing at test time. PPs are often compressed by the deduplication mechanism because the compressible PPs tend to be temporal and location adjuncts, which may be redundant across sentences. Without the manual deduplication mechanism, our model matches the ground truth around 80\% of the time. However, a low accuracy here may not actually cause a low final ROUGE score, as many compression choices only affect the final ROUGE score by a small amount. More details about compression options are in the Supplementary Material.



\section{Related Work}
\label{sec:rel}

\paragraph{Neural Extractive Summarization}
Neural networks have shown to be effective in extractive summarization. Past approaches have structured the decision either as binary classification over sentences \cite{Cheng_Neural_2016,Nallapati_SummaRuNNer_2017} or classification followed by ranking \cite{Narayan_Ranking_2018}. \citet{Zhou_Neural_2018} used a seq-to-seq decoder instead. For our model, text compression forms a module largely orthogonal to the extraction module, so additional improvements to extractive modeling might be expected to stack with our approach.

\paragraph{Syntactic Compression} Prior to the explosion of neural models for summarization, syntactic compression \citep{Martins_Summarization_2009,Woodsend_Learning_2011} was relatively more common.
Several systems explored the usage of constituency parses \cite{Berg-Kirkpatrick_Jointly_2011,Wang_A_2013,Li_Improving_2014} as well as RST-based approaches \cite{Hirao_Single_2013,Durrett_Learning_2016}. Our approach follows in this vein but could be combined with more sophisticated neural text compression methods as well. 

\paragraph{Neural Text Compression}
\citet{Filippova_Sentence_2015} presented an LSTM approach to deletion-based sentence compression. 
\citet{Miao_Language} proposed a deep generative model for text compression.
\citet{Zhang_Neural_2018} explored the compression module after the extraction model but the separation of these two modules hurt the performance. For this work, we find that relying on syntax gives us more easily understandable and controllable compression options.

Contemporaneously with our work, \citet{MendesEtAl2019} explored an extractive and compressive approach using compression integrated into a sequential decoding process; however, their approach does not leverage explicit syntax and makes several different model design choices.




\section{Conclusion}

In this work, we presented a neural network framework for extractive and compressive summarization. Our model consists of a sentence extraction model joined with a compression classifier that decides whether or not to delete syntax-derived compression options for each sentence. 
Training the model involves finding an oracle set of extraction and compression decision with high score, which we do through a combination of a beam search procedure and heuristics. Our model outperforms past work on the CNN/Daily Mail corpus in terms of ROUGE, achieves substantial gains over the extractive model, and appears to have acceptable grammaticality according to human evaluations.


\section*{Acknowledgments}

This work was partially supported by NSF Grant IIS-1814522, a Bloomberg Data Science Grant, and an equipment grant from NVIDIA. The authors acknowledge the Texas Advanced Computing Center (TACC) at The University of Texas at Austin for providing HPC resources used to conduct this research. Results presented in this paper were obtained using the Chameleon testbed supported by the National Science Foundation \cite{Keahey:19}.
Thanks as well to the anonymous reviewers for their helpful comments.

\bibliography{auxiliary_bib}
\bibliographystyle{acl_natbib}
\clearpage

\section*{Supplementary Material}

\appendix

\section{Experimental Setup}
\label{ap:exp}
\paragraph{Data Preprocessing}
We preprocess the datasets with the scripts provided by \citet{See_Get_2017}, which uses Stanford CoreNLP tokenization \citet{Manning_The_2014}. We use the non-anonymized version of the CNN/DM as in previous summarization work. 
For the New York Times Corpus, we filter out the examples with abstracts shorter than 50 words following the criteria in \cite{Durrett_Learning_2016}, yielding the NYT dataset. The statistics of the datasets are listed in Table~\ref{tb:data}. 
During sentence selection, we always select 3 sentences for CNN/DM and 5 sentences for NYT, which gave the best performance. For our syntactic analysis, all datasets are parsed with the constituency parser in Stanford CoreNLP \cite{Manning_The_2014}.

\begin{figure*}[h]
\centering
\includegraphics[width=0.7\textwidth]{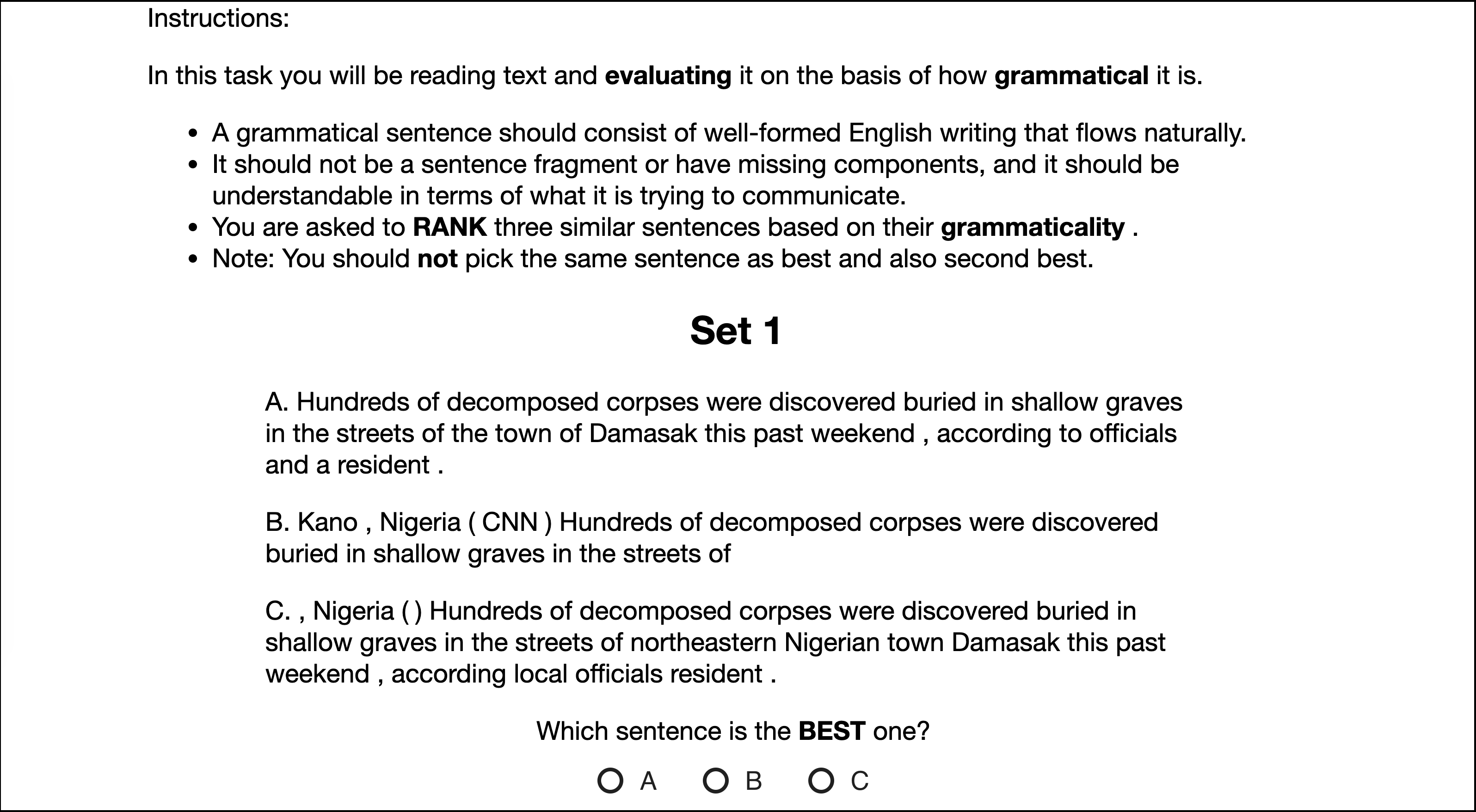}
\caption{The interface for Amazon turk human evaluation. All of the examples are fully shuffled. }
\label{pic:human}
\end{figure*}

\paragraph{Implementation Details}
We use the same pretrained word embeddings used in \cite{Narayan_Ranking_2018}. The size of the sentence and document representation vectors is 200. For the compression module, we use ELMo as the contextualized encoder without fine-tuning the parameter and project the vectors back to 200 dimensions after the ELMo layer. Dropout is applied after word embedding layers and LSTM layers at a rate of 0.2. We use the Adam optimizer \cite{Kingma_Adam_2014} with the initial learning rate at 0.001. The model converges after 2 epochs of training.
In initial experiments, we also found ELMo to be useful for sentence selection as well. However, to simplify comparisons with past work and due to scaling issues, we use it for compression only.
We use ROUGE \citep{Lin2004} for evaluation.\footnote{Command line parameters: ``-c 95 -m -n 2''}
During oracle construction, we use simplified unigram and bigram $\text{F}_1$ scores as a faster approximation to the full ROUGE.

\section{Turk Instructions}

\label{ap:human}
Figure~\ref{pic:human} shows the interface for Amazon turk human evaluation. 

\section{Type Analysis}
In Table~\ref{tb:surv-bf}, we show the statistics of the compression options in CNN. PP attachment and adjectives are the top 2 compression options and according to the oracle, more than half of PP and almost all of the adjectives are compressable without hurting the ROUGE.

\begin{table}[]
  \centering
  \small
  \renewcommand{\tabcolsep}{1.8mm}
  \begin{tabular}{c|ccccc}
  \toprule
  \multirow{1}{*}{Dataset} 
     & \#Train & \#Dev & \#Test & Ref len & Doc len  \\
\midrule
CNN  & 90266  & 1220 & 1093 & 37 & 540  \\
DM  & 196961  & 12148 & 10397 & 61 &  593 \\
NYT50  & 137778  & 17222 & 17223 & 88 &  727 \\
\bottomrule
  \end{tabular}
  \caption{
  Statistics of the CNN, Daily Mail, and NYT50 (see text) datasets. CNN features the shortest reference summaries overall, and this is where we find compression is most effective.
  }
  \label{tb:data}
\end{table}

\begin{table}[]
\centering
\small
\begin{tabular}{c|cccc}
\toprule
Node Type & \multicolumn{1}{l}{Len} & \multicolumn{1}{l}{\% of comps} & \multicolumn{1}{l}{Oracle comp \%}  \\
\midrule
PP        & 5.7                      & 39\%                          & 67\%                          \\
JJ        & 1.0                      & 19\%                          & 84\%                          \\
SBAR      & 12.1                      & 11\%                          & 59\%                          \\
ADVP      & 1.3                      & 7\%                           & 91\%                   \\\bottomrule      
\end{tabular}
\caption{Statistics of compression options in CNN. We show the top four constituency types that are compressible, along with the average length, the fraction of available compressions it accounts for, and how frequently the oracle says to compress these constituents.}
 \label{tb:surv-bf}
\end{table}

\end{document}